\documentclass[letterpaper]{article} 
\usepackage[preprint]{aaai2027}  

\makeatletter
\def\showauthors@on{T}
\makeatother

\usepackage[hyphens]{url}  
\usepackage{graphicx} 
\usepackage{natbib}  
\usepackage{caption} 
\usepackage{algorithm}
\usepackage{algorithmic}
\usepackage{amsmath}
\usepackage{multirow}
\usepackage[table]{xcolor}
\usepackage{enumitem}
\DeclareUnicodeCharacter{221E}{$\infty$}
\usepackage{newfloat}
\usepackage{listings}
\DeclareCaptionStyle{ruled}{labelfont=normalfont,labelsep=colon,strut=off} 
\floatstyle{ruled}
\newfloat{listing}{tb}{lst}{}
\floatname{listing}{Listing}

\usepackage{booktabs}

\title{A Sparse Glimpse of the Whole: Train-Free Self-Speculative Decoding}

\author{
    Yuesong Liu\textsuperscript{\rm 1}\equalcontrib,
    Yuan Zeng\textsuperscript{\rm 1}\equalcontrib,
    Min Lyu\textsuperscript{\rm 1}\corresponding,
    Ruilin Liu\textsuperscript{\rm 1},
    Yu Guo\textsuperscript{\rm 1},
    Yinlong Xu\textsuperscript{\rm 1}
}
\affiliations{
    \textsuperscript{\rm 1}University of Science and Technology of China
}

\begin{document}

\maketitle

\begin{abstract}
Speculative decoding alleviates the memory-bandwidth bottleneck in large language model inference, but its acceleration is jointly constrained by drafting overhead, token acceptance, and speculation length. We present a unified efficiency analysis showing that extending the speculation horizon can reduce rather than improve speedup when the marginal acceptance probability falls below the relative drafting cost. Guided by this analysis, we introduce \textbf{SparseSpec-L}, a training-free self-speculative decoding framework for long-context inference. SparseSpec-L generates lightweight drafts directly from the target model using a dynamically sparsified and recallable KV cache. It recycles per-head attention statistics produced during full-context verification as a no-extra-forward importance signal, allowing critical historical tokens to be recalled without permanently discarding the dense KV cache. An online entropy-based controller further selects the speculation length according to expected step-wise efficiency. Experiments across multiple long-context tasks and model scales show consistent end-to-end acceleration, with up to $2.79\times$ speedup over autoregressive decoding while preserving the target model's output distribution.
\end{abstract}


\section{Introduction}

\begin{figure*}[htp]
    \centering
    \includegraphics[width=0.9\textwidth]{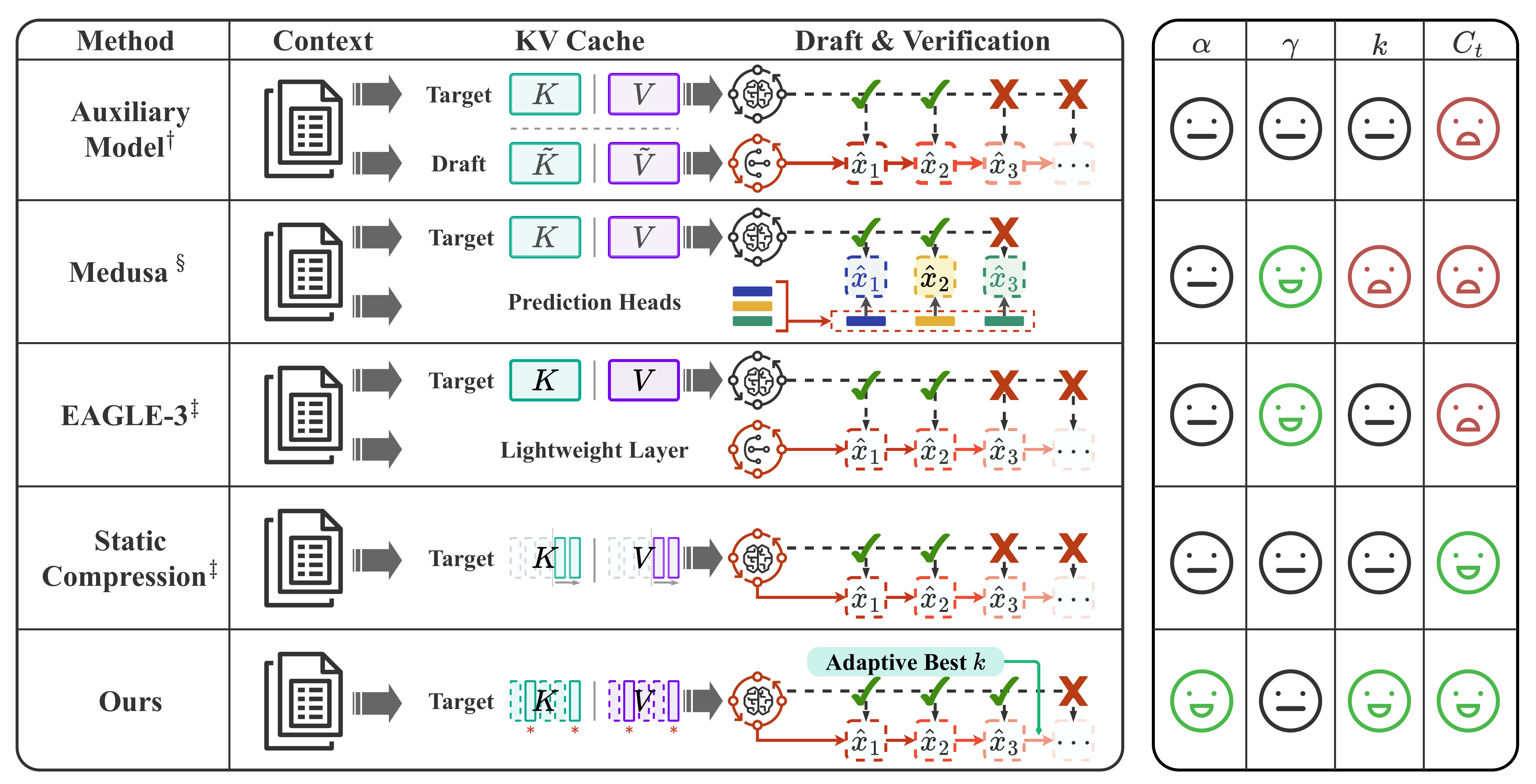}
    \caption{Comparison of existing speculative decoding paradigms ($\alpha$ is the token acceptance rate, $\gamma$ denotes drafting overhead, $k$ represents speculation length, and $C_t$ indicates retraining cost). Auxiliary models suffer from structural gaps that depress acceptance rates; Medusa fixes speculation length k at the architecture level; EAGLE-3 exhibits acceptance rate decay at large k; static compression permanently discards context, causing information loss.  Empirical evidence for each baseline's limitations is provided in: $^\dag$Tab.~\ref{tab:main_results}, $^\S$Appendix, and $^\ddag$Fig.~\ref{fig:eagle_and_snapkv}.}
    \label{fig:method_comparison}
\end{figure*}


Large language models (LLMs) demonstrate strong capabilities across
generation and reasoning tasks, but their autoregressive decoding
process imposes a substantial inference-latency bottleneck. Specifically, predicting each subsequent token requires loading the entire key-value (KV) cache of the preceding context into the processor, making decoding heavily memory-bandwidth bound~\cite{attentionAllYouNeed,pagedattention}. This challenge is further exacerbated in long-context scenarios, where linear memory growth and quadratic attention complexity~\cite{flashattention2} severely degrade real-time responsiveness. 

To mitigate this inefficiency, speculative decoding has emerged as a promising paradigm to break this bottleneck. By decoupling generation into a computationally cheap drafting stage and a parallel verification stage, it iteratively proposes and simultaneously verifies multiple candidate tokens. This effectively unlocks near-multiplicative speedups while mathematically guaranteeing the exact target distribution~\cite{unlocking}.

Despite its theoretical elegance, fully realizing the acceleration potential of speculative decoding remains a practical challenge. As categorized in Figure~\ref{fig:method_comparison}, representative approaches expose four common limitations:
(1) \textbf{Auxiliary Models}~\citep{acceleratingWithSpe,fastInference,specinfer} rely on an independent small drafter, but the inherent structural gap between the drafter and target model fundamentally depresses the acceptance rate. 
(2) \textbf{Prediction Heads} like Medusa~\citep{medusa} mitigate this gap by attaching drafting heads directly to the target model, but their speculation length $k$ is rigidly fixed at the architecture level, placing a hard mathematical ceiling on achievable speedups. 
(3) \textbf{Lightweight Layers} like Eagle-3~\citep{eagle3} enable flexible speculation lengths, but their auto-regressive approximation suffers a steep drop in acceptance rate for tail tokens, severely decaying performance at large $k$ and exhibiting extreme vulnerability to dataset dependency. 
(4) \textbf{Static Compression} methods~\citep{magicdec,RAPID} offer a training-free alternative by aggressively discarding historical context to minimize drafting overhead. However, this irreversible pruning permanently destroys context fidelity, causing persistent information loss and low acceptance rates in long-context scenarios. 
These representative paradigms expose a recurring trade-off among drafting overhead, speculation length, and acceptance rate. Although different methods occupy different operating points, none of the evaluated approaches simultaneously maintains low draft cost, robust long-context acceptance, and an adaptive speculation horizon.

To address these limitations, we introduce \textbf{SparseSpec-L}, a training-free self-speculative decoding framework for long-context inference. SparseSpec-L uses the target model itself as both drafter and verifier, avoiding the structural mismatch of an auxiliary draft model. During drafting, the target model accesses a dynamically sparsified KV cache, while verification retains the complete KV cache and therefore preserves the target distribution. Instead of permanently evicting historical context, SparseSpec-L recycles per-head attention statistics produced by the previous full-context verification pass to reconstruct a recallable sparse index for the next drafting round. This requires no additional model forward pass, although the current non-fused attention implementation still incurs kernel-level overhead. An online entropy-based controller further adjusts the speculation length according to real-time drafting confidence and measured drafting and verification costs.

Sparse-context target-model execution has also been explored by TriForce~\cite{triforce}, prior SparseSpec~\cite{pillaratten}, and Vegas~\cite{vegas}. SparseSpec-L shares their sparse-to-full principle but adopts a single-model, single-verification-stage pipeline, refreshes per-head indices from verification attention statistics, and selects the speculation horizon by expected step-wise efficiency. We therefore position it as an efficiency-oriented design for long-input workloads, rather than the first sparse-KV speculative decoding framework.

\textbf{Our contributions are threefold.}
(1) \textit{Efficiency analysis.} We formulate speculative decoding speedup as a joint function of drafting cost, prefix acceptance, and speculation length, and characterize the efficiency inversion that occurs when extending the drafting horizon no longer amortizes its marginal cost.
(2) \textit{Recallable sparse-context self-speculation.} We develop a training-free, single-model sparse-to-full decoding pipeline that refreshes per-head sparse KV indices using attention statistics from the preceding full-context verification pass, without permanently discarding historical KV states.
(3) \textit{Cost-aware adaptive speculation.} We estimate token-level acceptance from online entropy statistics and select the speculation length that maximizes expected step-wise efficiency. Experiments across long-context retrieval, reasoning, and synthesis tasks demonstrate consistent end-to-end acceleration of up to $2.79\times$.

\section{Rethinking Speculative Decoding}\label{sec:rethink}

To fundamentally understand the limitations of current acceleration algorithms and motivate the design of SparseSpec-L, we first abstract speculative decoding into a generalized framework. By quantifying the efficiency bottlenecks within this process, we can outline the exact blueprint required for an optimal solution.

\subsection{The Draft-Verification Paradigm}
Speculative decoding accelerates auto-regressive generation~\cite{autoregressiveLM} by decoupling the sequential process into a two-stage pipeline: (1) \textbf{Drafting stage.} A lightweight approximation mechanism generates $k$ candidate tokens, $\mathcal{X}_{\text{draft}}$ = \{$\hat{x}_{n+1},\ldots,\hat{x}_{n+k}$\}, aiming to mimic the target model's distribution with minimal compute. (2) \textbf{Verification stage.} The full target model evaluates $\mathcal{X}_{\text{draft}}$ in a single parallel forward pass. It identifies the longest prefix that satisfies the verification criteria: $\mathcal{X}_{\text{prefix}} = \{\hat{x}_{n+1},\ldots,\hat{x}_{n+m}\}$. The system then commits these $m$ tokens and generates an additional corrected token $x_{n+m+1}$. Consequently, each step produces $m+1$ tokens with only a single target model forward pass. This framework ensures mathematical equivalence to standard decoding while potentially providing multi-token speedup per step~\cite{acceleratingWithSpe,fastInference}.

\subsection{Unified Speedup Analysis}
\label{sec:efficiency}

To quantify the performance of speculative decoding, we define the speedup ratio $S$ as the throughput relative to standard decoding. Let $k$ denote the speculation length, which represents the number of candidate tokens generated by the drafting component in each step. Let $\alpha\in[0,1]$ be the expected acceptance rate, representing the expected proportion of draft tokens accepted per iteration ($\alpha = \mathbf{E}[m/k]$).
On average, the system produces $\alpha k + 1$ tokens in a single step.

The time cost of a parallel verification pass for $k$ tokens is approximately equivalent to that of a standard step, both of which we denote as $C_v$. Let $C_d$ be the per-token cost of the drafting component. The speedup $S$ can be formulated as:
\begin{equation}\label{eq:speedup}
S = \frac{(\alpha k + 1) C_v}{kC_d + C_v} = \frac{\alpha k + 1}{k(C_d/C_v) + 1}
\end{equation}

To understand the fundamental impact of the speculation length $k$, we first define the relative drafting overhead as $\gamma = C_d/C_v$. In an idealized scenario where the acceptance rate $\alpha$ is treated as a constant, the impact of $k$ on the speedup ratio $S$ can be quantified by its partial derivative:

\begin{equation}\label{eq:deriv}
\frac{\partial S}{\partial k} = \frac{\alpha - \gamma}{(k\gamma + 1)^2}
\end{equation}

Equation~\ref{eq:deriv} demonstrates that the benefit of extending the speculation length depends fundamentally on the magnitude of $\alpha$ relative to $\gamma$. If $\alpha > \gamma$, the speedup $S$ is a monotonically increasing function of $k$, suggesting that a longer speculation length always yields higher efficiency. Conversely, if $\alpha < \gamma$, the derivative $\frac{\partial S}{\partial k}$ remains negative, indicating that the drafting process is too costly relative to its accuracy and any speculation will degrade performance compared to standard decoding.


\begin{figure}[tbp]
    \centering
    \includegraphics[width=0.48\textwidth]{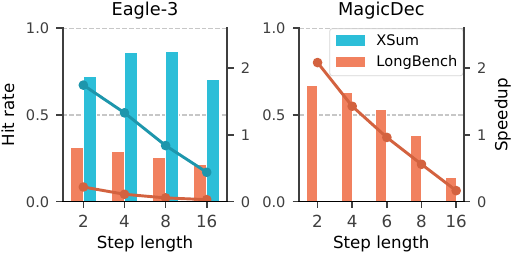}  
    \caption{Performance analysis of Eagle-3 and MagicDec(SnapKV) with variable speculation steps. For each method, the \textit{left y-axis} denotes token acceptance (hit) rate and the \textit{right y-axis} denotes speedup over auto-regressive decoding. Results are reported on LongBench~v2 and XSum with Llama3.1-8B-Instruct.}
    \label{fig:eagle_and_snapkv}
\end{figure}

While the constant $\alpha$ case provides a theoretical threshold, in practice, the expected acceptance rate $\alpha$ consistently decays as $k$ scales. As the speculation length increases, the compounding uncertainty of auto-regressive generation causes the acceptance rate for tail tokens to decay. Eventually, this triggers an \textit{efficiency inversion}, a tipping point where the marginal cost of drafting an additional token eclipses its expected contribution to the speedup ($\alpha_{\text{tail}} < \gamma$). Consequently, fixing $k$ across different contexts is inherently sub-optimal.

This theoretical inversion and the limitations of existing paradigms are empirically verified in Figure~\ref{fig:eagle_and_snapkv}. First, both training-based models and training-free static compression methods exhibit a severe speedup crater as $k$ increases, directly caused by the steep decay of $\alpha$. For instance, on LongBench v2, the speedup of MagicDec early peaks at $1.74\times$ when $k=2$. However, pushing the speculation length to $k=16$ triggers a catastrophic decline in $\alpha$ down to $6.3\%$, ultimately degrading the overall speedup to a sub-baseline $0.36\times$. Similarly, Eagle-3 sees its speedup regress from $2.24\times$ to $1.83\times$ on the XSum dataset under the same length extension. Second, Figure~\ref{fig:eagle_and_snapkv} exposes the critical dataset dependency of training-based methods. While Eagle-3 achieves a robust acceptance rate of approximately $67.2\%$ on the XSum dataset (commonly used in its original evaluation), its performance collapses drastically to a mere $8.4\%$ when shifted to the long-context LongBench scenario.

Therefore, maximizing acceleration is fundamentally a multidimensional optimization problem. As summarized in Figure~\ref{fig:method_comparison}, an ideal framework must simultaneously guarantee an acceptable drafting cost $\gamma$, a robust acceptance rate $\alpha$, and an optimal speculation length $k$. However, existing paradigms fail to achieve this. As demonstrated empirically, simply forcing a longer speculation length to chase higher speedups leads to a steep decline in $\alpha$, ultimately causing an efficiency inversion where actual acceleration drops. This acceleration bottleneck necessitates SparseSpec-L. Our approach pushes the practical speedup ceiling forward through two mechanisms: first, recallable self-speculation intrinsically guarantees a high and generalizable $\alpha$; second, an entropy-guided adaptive policy dynamically adjusts $k$ to constantly maintain generation within the peak efficiency zone, effectively mitigating the efficiency inversion.
\section{Methodology}
\label{sec:Methodology}

\begin{figure*}[t]
    \centering
    \includegraphics[width=\textwidth, height=8cm, keepaspectratio]{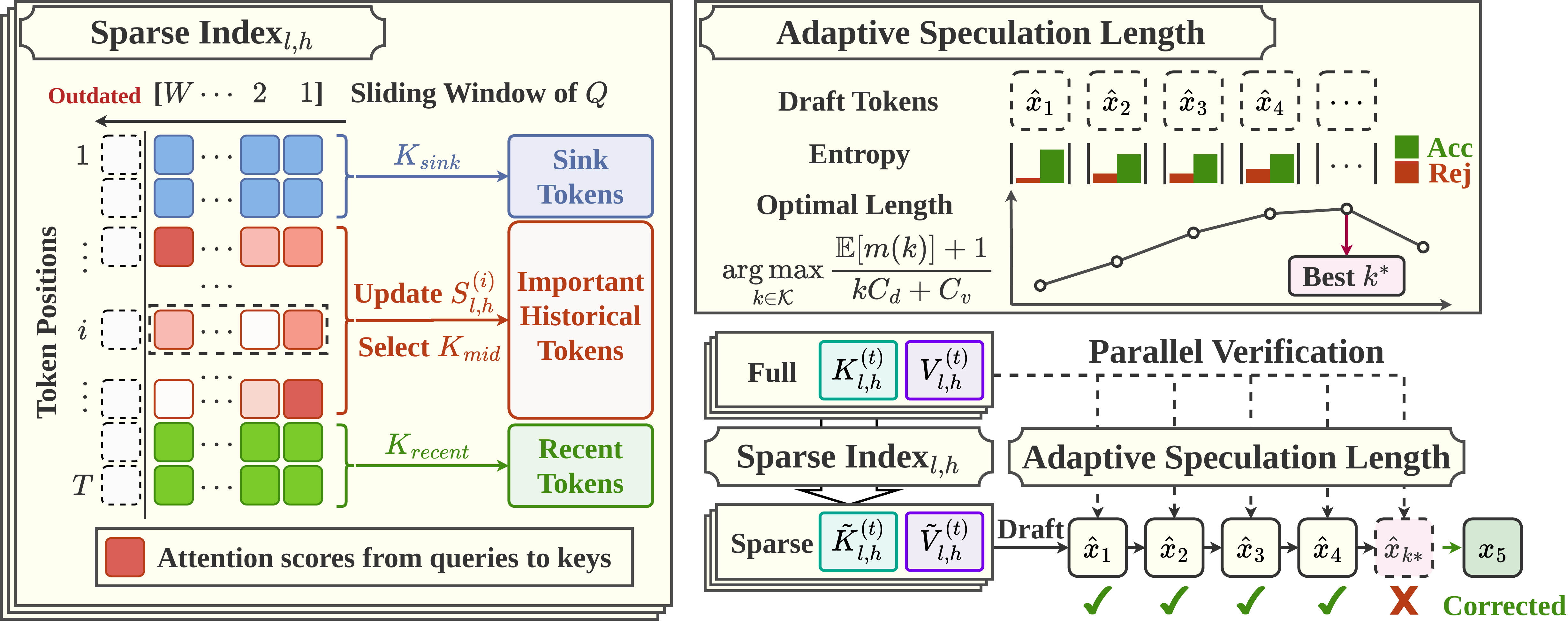}
    \caption{Overview of SparseSpec-L's sparse-to-full self-speculative decoding pipeline. \textit{Left}: the recallable sparse KV-cache index is constructed per attention head by retaining sink, important historical, and recent tokens, guided by attention scores from the previous verification pass. \textit{Right}: draft tokens are generated via sparse attention, then verified in a single parallel full-attention forward pass; accepted tokens are committed, attention weights refresh the sparse index, and the adaptive module selects the optimal $k^*$ based on per-token draft entropy.}
    \label{fig:overview}
\end{figure*}

Guided by the unified speedup analysis above, an ideal speculative decoding framework must concurrently maintain an acceptable drafting overhead $\gamma$, guarantee a robust and generalizable acceptance rate $\alpha$, and dynamically calibrate the speculation length $k$. To fulfill these criteria, we propose \textbf{SparseSpec-L}, a training-free, self-speculative decoding framework. It elegantly resolves these bottlenecks through two complementary mechanisms: recallable sparse-attention drafting to maximize $\alpha$ while bounding $\gamma$, and entropy-guided adaptive speculation to dynamically optimize $k$.

\subsection{Overview of SparseSpec-L}
SparseSpec-L fundamentally adopts a sparse-to-full self-speculative framework, wherein a single target model acts as both the drafter and the verifier, inherently eliminating any structural mismatch. Figure~\ref{fig:overview} illustrates the two-stage workflow executed at every decoding iteration.

\textbf{Sparse Drafting Stage.} The target model autoregressively generates $k$ speculative tokens using a recallable sparse KV cache. Instead of static pruning, the retained KV positions are dynamically guided by importance scores inherited from the previous verification pass, ensuring the drafter attends only to the most context-critical subset of history.

\textbf{Full Verification Stage.} The $k$ draft tokens are verified in one parallel forward pass using the complete KV cache. In our current implementation, verification uses dense attention that materializes per-head attention statistics. SparseSpec-L recycles these statistics to refresh token importance for the next drafting round. This introduces no additional model forward pass, but the use of non-fused dense attention incurs implementation overhead relative to an optimized FlashAttention-based verifier.

We next describe the recallable sparse-index construction and the entropy-guided policy for adaptive speculation lengths.

\subsection{Sparse Attention Drafting}\label{sec:sparse_drafting}

\textbf{Attention-Based Importance Estimation.}
We estimate the importance of historical tokens from the query--key attention statistics materialized during full-context verification. The statistics are reused rather than recomputed through an additional forward pass. Importance is maintained independently for every layer $l$ and attention head $h$, allowing the sparse draft cache to preserve heterogeneous head-specific retrieval patterns~\cite{attentionhead}.

For each token position $i \in \{1, \ldots, T\}$, its importance score $S^{(i)}_{l,h}$ is calculated by aggregating the attention it received from the most recent $W$ query positions within that specific head:
\begin{equation}
S^{(i)}_{l,h} = \sum_{q=T-W+1}^{T} a^{(q \to i)}_{l,h}
\label{eq:score}
\end{equation}
where $a^{(q \to i)}_{l,h}$ denotes the attention weight from query $q$ to key $i$, and the sliding window $W$ effectively controls the recency bias of the signal.

\textbf{Recallable Index Construction.} 
Based on $S^{(i)}_{l,h}$, we independently construct a sparse index set $\mathcal{I}_{l,h}$ for each head. To ensure comprehensive context fidelity under a total budget $K_{\text{total}}$, each index set retains three complementary groups of KV positions: (1) \textbf{Sink tokens:} the first $K_{\text{sink}}$ positions, encoding essential global context~\citep{streamingLLM}; (2) \textbf{Recent tokens:} the most recent $K_{\text{recent}}$ positions, supplying indispensable local context for autoregression; and (3) \textbf{Important historical tokens:} the top $K_{\text{mid}}$ positions ranked by $S^{(i)}_{l,h}$, drawn exclusively from the remaining non-sink, non-recent span to capture critical long-range dependencies. Formally, the retained index set is defined as:
\begin{equation}
\begin{split}
\mathcal{I}_{l,h} = & \{1, \ldots, K_{\text{sink}}\} \cup \text{TopK}_{K_{\text{mid}}}(S_{l,h}) \\
& \cup \{T-K_{\text{recent}}+1, \ldots, T\}
\end{split}
\label{eq:index}
\end{equation}
where $\text{TopK}$ excludes sink and recent positions. The sparse KV cache used for drafting is then instantiated via a low-overhead index gathering:
\begin{equation}
\begin{split}
\tilde{K}_{l,h} &= \text{Gather}(K_{l,h}, \mathcal{I}_{l,h}), \\
\tilde{V}_{l,h} &= \text{Gather}(V_{l,h}, \mathcal{I}_{l,h})
\end{split}
\label{eq:gather}
\end{equation}
Because the original dense KV tensors remain resident, the sparse index $\mathcal{I}_{l,h}$ can be reconstructed after every verification step without irreversible KV eviction.

\textbf{Hyperparameter Selection.} 
The total budget $K_{\text{total}} = K_{\text{sink}} + K_{\text{mid}} + K_{\text{recent}}$ directly dictates the drafting compression ratio. In our implementation, $K_{\text{sink}}$ and $K_{\text{recent}}$ follow established empirical constants that demonstrate robustness across model families~\citep{streamingLLM, snapkv}, while $K_{\text{mid}}$ dynamically fills the remaining budget. The aggregation window is fixed at $W=16$, which empirically yields stable importance signals across all evaluations.

\subsection{Adaptive Speculation Length}\label{sec:adaptive_len}

To avoid the efficiency inversion caused by a rigid speculation length, SparseSpec-L employs a lightweight, training-free policy that dynamically modulates $k$ based on the drafter's real-time confidence.

\textbf{Entropy-Based Acceptance Prediction.} 
During the sparse drafting stage, we actively record the output entropy $H_i$ of each generated candidate token. Following the subsequent verification step, every drafted token is categorically labeled as accepted or rejected. This allows us to track the empirical mean entropy of the accepted class ($\bar{H}_{\text{acc}}$) and the rejected class ($\bar{H}_{\text{rej}}$) via exponential moving averages with a smoothing coefficient $\beta$:
\begin{equation}
\begin{split}
\bar{H}_{\text{acc}} &\gets \beta \bar{H}_{\text{acc}} + (1 - \beta) H_{\text{acc}}, \\
\bar{H}_{\text{rej}} &\gets \beta \bar{H}_{\text{rej}} + (1 - \beta) H_{\text{rej}}
\end{split}
\label{eq:ema}
\end{equation}

For any newly drafted token $i$ with entropy $H_i$, we estimate its soft acceptance probability $p_i$ by evaluating its proximity to these tracked class centers. We apply a normalized softmax over their negative L1 distances:
\begin{equation}
p_i = \frac{e^{-|H_i - \bar{H}_{\text{acc}}|}}{e^{-|H_i - \bar{H}_{\text{acc}}|} + e^{-|H_i - \bar{H}_{\text{rej}}|}}
\label{eq:prob}
\end{equation}

\textbf{Optimal Length Selection.} 
Because speculative decoding enforces a strict prefix-matching criterion, the probability that the first $m$ drafted tokens are consecutively accepted is given by the joint probability $\prod_{i=1}^{m} p_i$, assuming conditional independence. Therefore, for a given speculation length $k$, the expected number of accepted draft tokens is the sum of these prefix probabilities: $\sum_{m=1}^{k} \prod_{i=1}^{m} p_i$.

\begin{table*}[tp]
\centering
\footnotesize
\caption{End-to-end results on LongBench v2 and $\infty$-Bench ($\alpha$: acceptance rate, Aver.$k$: average accepted tokens, T: throughput (tokens/s), S: speedup over autoregressive decoding). Auxiliary Model is not applicable to Qwen2.5-7B-Instruct due to vocabulary mismatch. Eagle-3 uses the official released checkpoint which lacks Qwen support.}\label{tab:main_results}
\begin{tabular}{lcccccccc}
\toprule
\multirow{2}{*}{\textbf{Method}} 
& \multicolumn{4}{c}{\textbf{LongBench}} 
& \multicolumn{4}{c}{\textbf{$\infty$-Bench}} \\
\cmidrule(lr){2-5} \cmidrule(lr){6-9} 
& $\alpha$ & Aver.$k$ & T & S 
& $\alpha$ & Aver.$k$ & T & S \\

\midrule
\multicolumn{9}{c}{\textbf{\textit{Llama-3.1-8B-Instruct}}} \\
Autoregressive & - & 1 & 3.71 & 1.0$\times$
               & - & 1 & 3.94 & 1.0$\times$\\

Auxiliary Model   & 50.3\% & 2.94 & 4.53 & 1.22$\times$
            & 52\% & 1.78 & 3.78 & 0.95$\times$ \\

MagicDec (StreamLLM)    & 33.5\% & 3.59 & 5.4 & 1.45$\times$
            & 50.9\% & 5.30 & 7.65 & 1.94$\times$ \\
MagicDec (SnapKV)    & 80.1\% & 1.59 & 6.46 & 1.74$\times$
            & 83.4\% & 1.66 & 6.62 & 1.68$\times$ \\

RAPID & 27.9\% & 2.73 & 5.34 & 1.43$\times$
      & 29.1\% & 2.73 & 5.42 & 1.37$\times$ \\

Eagle-3 & 1.95\% & 1.07 & 3.30 & 0.89$\times$
      & 2.78\% & 1.11 & 3.62 & 0.92$\times$ \\

SparseSpec-L (Ours) & 72.9\% & 11.26 & 10.38 & \textbf{2.79$\times$} 
              & 79.4\% & 8.41 & 10.25 & \textbf{2.60$\times$} \\

\midrule  
\midrule
\multicolumn{9}{c}{\textbf{\textit{Qwen2.5-7B-Instruct}}} \\  
Autoregressive & - & 1 & 5.24 & 1.0$\times$
               & - & 1 & 5.24 & 1.0$\times$\\

MagicDec (StreamLLM)    & 28.2\% & 3.03 & 5.92 & 1.12$\times$ 
            & 28.8\% & 3.06 & 5.95 & 1.13$\times$ \\ 

MagicDec (SnapKV)   & 65.9\% & 1.32 & 7.49 & 1.42$\times$  
            & 67.4\% & 1.34 & 7.55 & 1.44$\times$ \\

RAPID & 27.9\% & 1.36 & 5.67 & 1.08$\times$   
            & 24.7\% & 1.27 & 5.37 & 1.02$\times$ \\ 

SparseSpec-L (Ours) & 58.8\% & 8.03 & 10.24 & \textbf{1.95$\times$}
            & 52\% & 8.4 & 9.77 & \textbf{1.86$\times$} \\

\bottomrule
\end{tabular}
\label{tab:main}
\end{table*}

By adding the single correction token guaranteed during the verification pass and substituting this expected total length back into our unified speedup formulation (Eq.~\ref{eq:speedup}), we dynamically select the optimal speculation length $k^*$ from a pre-defined candidate set $\mathcal{K}$ that strictly maximizes the expected step-wise efficiency:
\begin{equation}
k^* = \arg\max_{k \in \mathcal{K}} \frac{1 + \sum_{m=1}^{k} \prod_{i=1}^{m} p_i}{k C_d + C_v}
\label{eq:optimal_k}
\end{equation}
where $C_d$ and $C_v$ represent the drafting and verification cost, respectively.

This dynamic modulation acts as an online feedback controller. When generation is confident (low entropy, yielding high $p_i$), the policy aggressively extends $k^*$ to harvest more accepted tokens per step. Conversely, when uncertainty spikes, the policy preemptively contracts $k^*$ to avoid costly rejections. Consequently, SparseSpec-L seamlessly transforms the notorious efficiency inversion from a hard bottleneck into a dynamically managed operating point, optimizing acceleration throughout the decoding process.

\section{Experiments}\label{sec:exp}

\subsection{Experimental Setup}\label{sec:expsetup}

\paragraph{Models \& Datasets.} 
Our primary controlled comparison uses Llama-3.1-8B-Instruct (L3) and Qwen2.5-7B-Instruct (Q2) on LongBench v2~\cite{longbenchv2} and InfiniteBench~\cite{infinitebench}. To evaluate generalization across model scales and task types, we additionally evaluate Llama-3.2-3B-Instruct and Qwen2.5-14B-Instruct on LongBench v2, RULER~\cite{ruler} QA1, RULER NIAH, and InfiniteBench. These workloads cover multi-task reasoning and synthesis, long-context question answering, and needle retrieval.

\paragraph{Baselines.} 
We compare against an auxiliary drafter~\cite{specinfer}, EAGLE-3~\cite{eagle3}, LayerSkip, MagicDec with StreamingLLM or SnapKV~\cite{magicdec,streamingLLM,snapkv}, and RAPID~\cite{RAPID}. We use official checkpoints and recommended configurations when available, and report each fixed-length baseline under its best-performing speculation length. Medusa~\cite{medusa} is omitted from the main table because no aligned checkpoint is available for our target models; an evaluation using its released Vicuna-7B checkpoint is provided in the supplementary material.  

\paragraph{Implementation Details.} 
All primary experiments are conducted on a single NVIDIA A40 GPU with 48\,GB memory using Hugging Face Transformers. Prefill is identical for SparseSpec-L and the autoregressive baseline; SparseSpec-L modifies only the decoding phase and therefore does not change time to first token under the same prefill implementation. To materialize per-head attention statistics, the current verifier uses dense non-FlashAttention kernels. These statistics require no additional forward pass, but the non-fused implementation increases verification latency relative to an optimized autoregressive attention kernel. The sparse-index update costs 41.1\,ms per decoding iteration in the primary 60K setting, while the entropy controller contributes less than 0.5\% of total step latency. Unless otherwise specified, the context length is capped at 64K and the generation length at 128 tokens.

\subsection{Main Results}
\label{sec:main_results}

Table~\ref{tab:main_results} reports the performance of all baselines on L3 and Q2 across two benchmarks. Note that while SparseSpec-L utilizes an adaptive speculation length, all other baselines are reported under their optimal fixed speculation length that yields the highest speedup. Due to hardware memory constraints, we strictly cap the maximum context length at 64K tokens and the maximum generation length at 128 tokens across all evaluations.

\textbf{Generalization across models and tasks.}
Table~\ref{tab:generalization} extends the evaluation beyond the two primary configurations. SparseSpec-L provides $1.55\times$--$2.79\times$ speedup across 3B--14B models and across multi-task reasoning, question answering, needle retrieval, and long-context synthesis. Acceptance remains between 72.9\% and 84.6\% in these settings. These results indicate that the acceleration is not restricted to a single model size or benchmark.

\textbf{Comparison with layer-skipping self-speculation.}
We additionally evaluate LayerSkip using its official configuration. It achieves $0.80\times$ speedup with 41.4\% acceptance on normal-length inputs and $0.77\times$ with 38.4\% acceptance on the 60K LongBench setting. Under this evaluated configuration, the limited number of accepted draft tokens is insufficient to amortize the verification cost. We report the complete setup and throughput results in the supplementary material.

\textbf{Batch scaling.}
At 16K context, SparseSpec-L achieves $1.10\times$, $1.52\times$, and $2.45\times$ speedup at batch sizes 1, 2, and 4, respectively, compared with $0.91\times$, $1.24\times$, and $1.92\times$ for MagicDec. Its relative throughput advantage therefore increases from $1.21\times$ to $1.28\times$. Full throughput, acceptance, and memory statistics are provided in the supplementary material.

\begin{table}[t]
\centering
\small
\setlength{\tabcolsep}{3.5pt}
\caption{Cross-model and cross-task generalization of SparseSpec-L.
Speedup is measured against autoregressive decoding under the same
hardware and sequence configuration.}
\label{tab:generalization}
\begin{tabular}{llccc}
\toprule
Model & Dataset & Acc. & Aver.$k$ & Speedup \\
\midrule
Llama-3.2-3B & LongBench v2 & 82.0\% & 5.60 & 1.91$\times$ \\
Llama-3.2-3B & RULER QA1 & 76.2\% & 5.88 & 1.63$\times$ \\
Llama-3.2-3B & RULER NIAH & 83.0\% & 7.99 & 1.96$\times$ \\
Llama-3.2-3B & InfiniteBench & 77.0\% & 7.24 & 2.41$\times$ \\
Llama-3.1-8B & LongBench v2 & 72.9\% & 11.26 & 2.79$\times$ \\
Qwen2.5-14B & LongBench v2 & 84.6\% & 5.66 & 1.55$\times$ \\
\bottomrule
\end{tabular}
\end{table}

\begin{table}[t]
\centering
\footnotesize
\setlength{\tabcolsep}{4pt}
\caption{Ablation on speculation length.}
\label{tab:ablation_k}
\begin{tabular}{lcccc}
\toprule
\textbf{Method} & $\alpha$ & Aver.$k$ & $T$ & $S$ \\
\midrule
Fixed $k=4$  & 93.4\% & 3.66  & 7.81  & 2.10$\times$ \\
Fixed $k=8$  & 86.7\% & 6.75  & 8.89  & 2.39$\times$ \\
Fixed $k=12$ & 78.5\% & 9.16  & 9.99  & 2.69$\times$ \\
Fixed $k=16$ & 72.6\% & 11.31 & 9.90  & 2.66$\times$ \\
Fixed $k=20$ & 67.9\% & 13.01 & 9.75  & 2.62$\times$ \\
\midrule
\textbf{Adaptive} & \textbf{72.9\%} & \textbf{11.26} & \textbf{10.38} & \textbf{2.79$\times$} \\
\bottomrule
\end{tabular}
\end{table}

\textbf{Primary comparison.}
SparseSpec-L achieves $2.79\times$/$2.60\times$ speedup on
LongBench v2/InfiniteBench with L3, and
$1.95\times$/$1.86\times$ with Q2. On LongBench v2, it combines
$72.9\%$ acceptance with 11.26 accepted draft tokens per iteration.
In comparison, MagicDec--SnapKV reaches $80.1\%$ acceptance only under
the short fixed horizon $k=2$, yielding 1.59 accepted draft tokens and
$1.74\times$ speedup.

The auxiliary drafter attains a lower acceptance rate than SparseSpec-L
in the same setting (50.3\% versus 72.9\%). The official EAGLE-3
checkpoint performs well on its in-domain XSum setting but transfers
poorly to 60K LongBench v2, where acceptance falls below 3\%.
SparseSpec-L also maintains $1.95\times$ acceleration on Q2 without
model-specific training.


\begin{figure}[t]
    \centering
    \includegraphics[width=\linewidth]{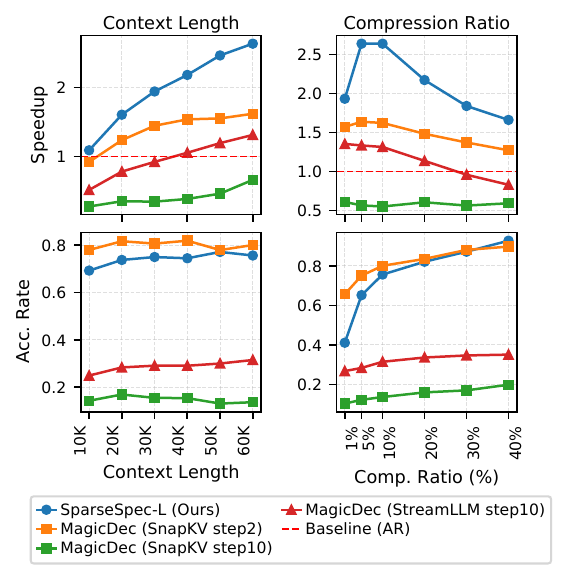}
    \caption{Sensitivity analysis on LongBench v2 with Llama-3.1-8B-Instruct model. The performance (Top: Speedup; Bottom: Acceptance rate) are evaluated against two critical dimensions: context length (Left) and KV-cache compression ratio (Right).}
    \label{fig:sensitivity_analysis}
\end{figure}

\subsection{Ablation Study and Sensitivity Analysis}\label{sec:ablation}

\paragraph{Contribution of Adaptive Speculation Policy.} Table~\ref{tab:ablation_k} compares our entropy-guided adaptive policy against fixed speculation lengths ($k \in \{4, 8, 12, 16, 20\}$) on L3 (LongBench v2). Pushing a fixed horizon inevitably degrades acceptance rate (93.4\% at $k{=}4$ to 67.9\% at $k{=}20$), empirically confirming the efficiency inversion in Eq.~\ref{eq:speedup}: throughput peaks at $k{=}12$ (2.69$\times$) then falls. Our adaptive policy avoids this inversion and outperforms all fixed-$k$ configurations without manual tuning, reaching $2.79\times$ speedup. Per-example speedup distributions are reported in the supplementary material.

The benefit of adaptivity depends on workload heterogeneity. On RULER NIAH, adaptive control improves token acceptance from 68.6\% to 79.9\% and throughput from 17.11 to 20.07 tokens/s relative to fixed $k=10$, corresponding to a 17.3\% throughput improvement. On InfiniteBench, adaptive control closely matches the best fixed setting (14.81 versus 14.75 tokens/s), while avoiding task-specific manual selection of $k$. Thus, adaptivity is particularly useful when token difficulty varies throughout generation, but may provide little additional gain on more uniform workloads.

As a sanity check, negative draft entropy obtains an acceptance AUC of 0.638 on RULER QA1. Tokens in the low-entropy bin are accepted at 86.3\%, compared with 62.1\% in the high-entropy bin. This modest but consistent separation supports entropy as a useful online control signal; it does not imply that entropy alone is a fully calibrated acceptance predictor.


\paragraph{Scaling with Context Length.} 
Figure~\ref{fig:sensitivity_analysis} traces performance as context length scales from 10K to 60K tokens. While all methods exhibit stable acceptance rates and increasing speedups at longer contexts, SparseSpec-L consistently achieves the highest acceleration. Crucially, the competitive acceptance rate of MagicDec (SnapKV) is strictly conditioned on its optimal fixed speculation length of $k=2$. When forced to a longer horizon ($k=10$), its acceptance rate degrades below that of StreamingLLM due to permanent information loss. In contrast, SparseSpec-L robustly sustains both a high acceptance rate and a large effective speculation length across all context lengths. 

\paragraph{Sensitivity to Compression Ratio.} 
Modulating the KV-cache compression ratio exposes the fundamental trade-off between drafting overhead ($\gamma$) and acceptance rate ($\alpha$). Retaining more KV caches increases $\alpha$ but incurs higher drafting costs, eventually degrading the overall speedup. As shown in Figure~\ref{fig:sensitivity_analysis}, SparseSpec-L consistently dominates all baselines in both speedup and acceptance rate across various compression regimes (with the sole exception of MagicDec (SnapKV) operating at a heavily restricted $k=2$). The empirical optimum lies at a compression ratio of ${\sim}10\%$, balancing context fidelity and drafting overhead.

\section{Related Work}
\label{sec:related}

\textbf{Speculative decoding.}
Speculative decoding accelerates autoregressive generation through cheap drafting and parallel verification~\cite{acceleratingWithSpe,fastInference,specinfer}. Auxiliary drafters incur additional memory and model mismatch, while Medusa, multi-token prediction, EAGLE-3, and layer-skipping methods require model-specific parameters or sacrifice representational capacity~\cite{medusa,MTP,eagle3}.

\paragraph{KV cache compression.}
StreamingLLM, H2O, ScissorHands, Quest, and PyramidKV reduce KV access through eviction or sparse selection~\cite{streamingLLM,h2o,scissorhands,quest,pyramidkv}. MagicDec uses compressed KV caches for speculative decoding, while InfiniGen~\cite{infinigen} and ArkVale~\cite{arkvale} support recallable KV recovery. SparseSpec-L retains dense KV states for verification and uses a recallable sparse view only for drafting.

\textbf{Sparse-context and adaptive speculation.}
TriForce uses hierarchical sparse-KV target execution before full-KV verification, while prior SparseSpec and Vegas also perform sparse-context self-speculation~\cite{triforce,pillaratten,vegas}. SparseSpec-L differs in its single-stage pipeline, per-head verification-guided index, and cost-aware length objective. Dynamic draft lengths have also been studied by DISCO, AdaEDL, and SVIP~\cite{disco,adaedl,svip}; our controller directly maximizes estimated step efficiency using online entropy and measured latency.
\section{Conclusion}
\label{sec:conclusion}

We introduced SparseSpec-L, a training-free sparse-to-full self-speculative framework for long-context inference. It combines a recallable per-head sparse KV index with cost-aware adaptive speculation, achieving up to $2.79\times$ end-to-end speedup across multiple tasks and model scales while preserving full-KV verification.

\section*{Limitations}
Our implementation uses non-fused sparse-KV gathering and dense non-FlashAttention verification, leaving room for fused kernel optimization. Experiments are limited to one A40, contexts up to 64K, and 128 generated tokens. Evaluation on optimized serving engines and longer outputs remains future work. The entropy controller is only moderately predictive and provides task-dependent gains.

\bibliography{aaai2027}



\end{document}